\documentclass{article}
\usepackage{spconf,amsmath,graphicx}
\usepackage{comment}
\usepackage{xhfill}
\usepackage{adjustbox}
\usepackage{arydshln}
\usepackage{adjustbox}
\usepackage[T1]{fontenc}% http://ctan.org/pkg/fontenc
\graphicspath{ {images/} }

\title{Deep Multi-view Models for Glitch Classification}
\name{Sara Bahaadini$^{\star}$, Neda Rohani$^{\star}$, Scott Coughlin$^{\star \star}$, Michael Zevin$^{\star \star}$, Vicky Kalogera$^{\star \star}$, Aggelos K Katsaggelos$^{\star}$}
\address{$^{\star}$ Department of Electrical Engineering and Computer Science, Northwestern University, Evanston, IL, USA\\
$^{\star \star}$CIERA Physics and Astronomy, Northwestern University, Evanston, IL, USA
}
\begin{document}
%\ninept
%
\maketitle
%

% Comment this line out
                                                          % if you need a4paper
%\documentclass[a4paper, 10pt, conference]{ieeeconf}      % Use this line for a4
                                                          % paper

% The following packages can be found on http:\\www.ctan.org
%\usepackage{graphics} % for pdf, bitmapped graphics files
%\usepackage{epsfig} % for postscript graphics files
%\usepackage{mathptmx} % assumes new font selection scheme installed
%\usepackage{times} % assumes new font selection scheme installed
%\usepackage{amsmath} % assumes amsmath package installed
%\usepackage{amssymb}  % assumes amsmath package installed

\thispagestyle{empty}
\pagestyle{empty}

%%%%%%%%%%%%%%%%%%%%%%%%%%%%%%%%%%%%%%%%%%%%%%%%%%%%%%%%%%%%%%%%%%%%%%%%%%%%%%%%
\begin{abstract}
Non-cosmic, non-Gaussian disturbances known as ``glitches'', show up in gravitational-wave data of the Advanced Laser Interferometer Gravitational-wave Observatory, or aLIGO. In this paper, we propose a deep multi-view convolutional neural network to classify glitches automatically. The primary purpose of classifying glitches is to understand their characteristics and origin, which facilitates their removal from the data or from the detector entirely.
We visualize glitches as spectrograms and leverage the state-of-the-art image classification techniques in our model. The suggested classifier is a multi-view deep neural network that exploits four different views for classification. The experimental results demonstrate that the proposed model improves the overall accuracy of the classification compared to traditional single view algorithms.
\end{abstract}

\begin{keywords}
Multi-view learning, deep learning, image classification, neural network
\end{keywords}
%%%%%%%%%%%%%%%%%%%%%%%%%%%%%%%%%%%%%%%%%%%%%%%%%%%%%%%%%%%%%%%%%%%%%%%%%%%%%%%%
\section{INTRODUCTION}

In many machine learning problems, samples are collected from more than one source. Also, various feature extraction methods can be used to provide more than one set of feature vectors per sample. Such extra sources or feature vectors are referred to as ``views". Using multiple views can improve performance as they may provide complementary or redundant information \cite{xu2013survey}.

Fusion of multiple sources of information has been used in many applications such as emotion recognition \cite{huang2016multi}, recommendation systems \cite{vahidrecomm}, speech recognition \cite{saraieee,bahaadini2014posterior}, and biometric verification \cite{aleksic2003audio}.
Integrating multiple sources of data is a challenging task, and various approaches have been proposed in the literature \cite{saraieee}\cite{7362367}. More recently, deep learning techniques have shown promising performance for multi-modal fusion \cite{srivastava2012multimodal,ngiam2011multimodal,suk2014hierarchical}. Moreover, deep learning methods have shown superb performance for many classification problems including image classification. In this paper, we propose deep multi-view models for a particular classification problem from the aLIGO project \cite{jointpaper}. %\textcolor{blue}{I think you can just do what Kevin did here and cite the paper in prep, then change it once it is published.}.

Advanced LIGO (Advanced Laser Interferometer Gravitational -wave Observatory, or aLIGO) has recently made the first direct observations of gravitational waves \cite{GW150914} \cite{GW151226}, which are ripples in the fabric of spacetime caused by accelerating masses. Since aLIGO is sensitive to minuscule changes in distance, its experimental data is affected by a variety of non-cosmic disturbances. When such disturbances, called ``glitches'', show up in the gravitational-wave data, they generally worsen the quality of the detection of candidate cosmic signals. The elimination and prevention of these glitches will improve the quality of the detection system and increase the chance of detecting gravitational waves. Therefore, it is necessary to develop methods for identifying and characterizing glitches, which will help to determine their origin and eliminate their cause. Since glitches can be visualized as time-frequency-energy images (spectrograms), image classification techniques can be used to identify and characterize them. 
\begin{figure}[t!]
\centering
\includegraphics[width=0.4\textwidth]{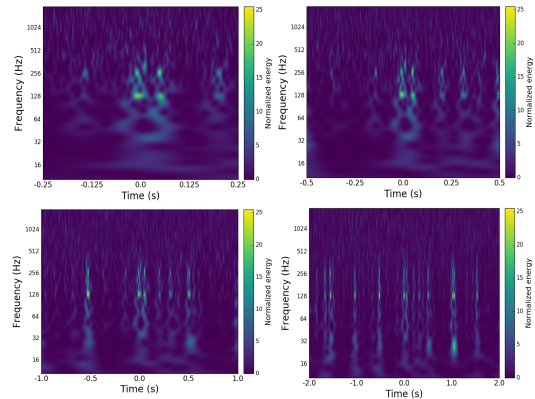}
\caption{An example of a Helix glitch with four views (Top from left to right: $0.5$ and $1.0$ seconds, bottom from left to right: $2.0$ and $4.0$ seconds).}
\label{fig:ex}
\end{figure}
\begin{figure*}[th!]
\centering
\includegraphics[width=18cm]{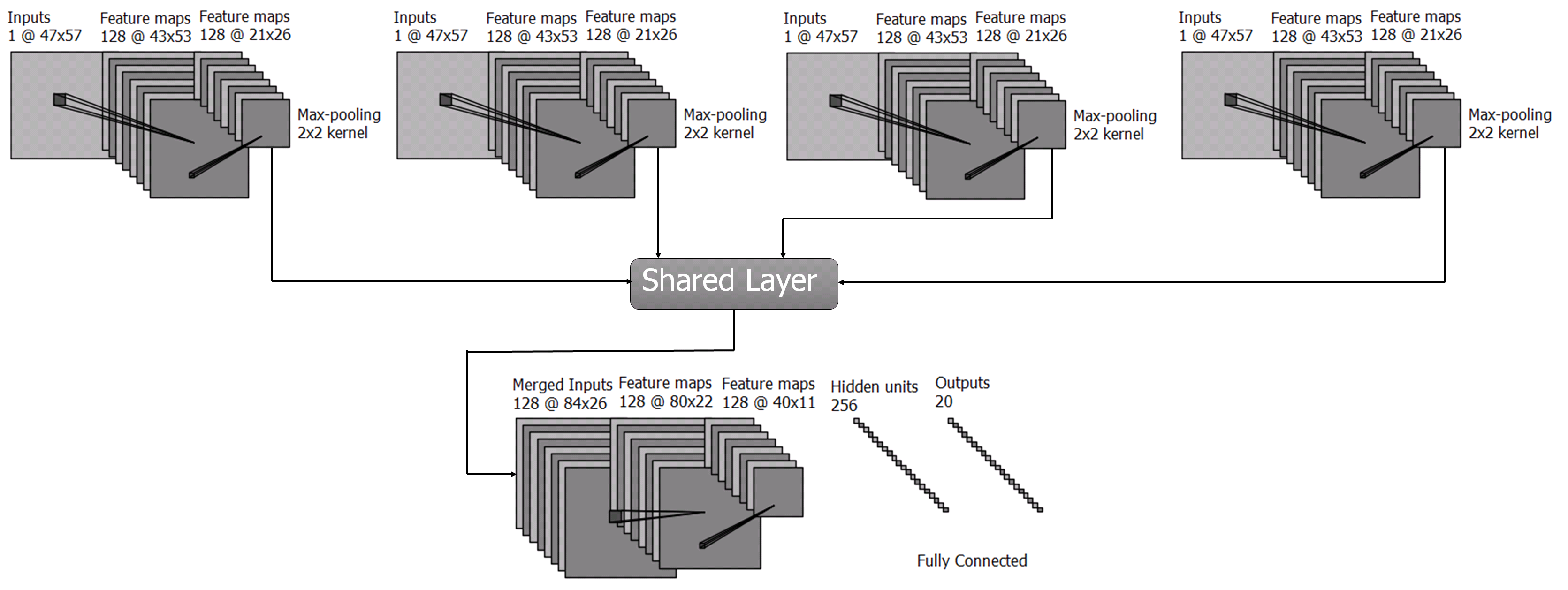}
\caption{Schematic representation of the ``parallel view" model.}
\label{fig:pview}

\end{figure*}

In this paper, we present the development of a multi-view deep neural network framework for glitch classification. Compared to standard methods that use just one set of images, we propose  four-input models.
We exploit four different time durations that are available for each glitch, namely each glitch plotted over time windows of $0.5$, $1$, $2$, and $4$ seconds. An example of such a glitch (from the ``Helix'' class) with four durations is shown in Fig.~\ref{fig:ex}. We suggest two multi-view deep neural network models (a parallel view and a merged view) and we compare their performances to deep single view models. Experimental analysis shows that single view models trained with shorter glitches have a better performance for the classes that have shorter duration, while single view models trained with longer-duration glitches work better for long-duration classes.
Our experimental results show that the developed multi-view framework improves the glitch classification accuracy by capturing the required information from glitches of various morphological characteristics.

The rest of this paper is organized as follows. In the next section, we present our model. Experiments and results are discussed in Section \ref{sec:exp}. We conclude this paper in Section \ref{sec:conc}.
%\vspace{-0.3 cm}
\section{THE PROPOSED MODEL}
\label{sec:model}
%We have used different types of layers, which are common in deep learning especially in convolutional neural networks (CNNs), for our models. In the following, we first explain layers' types in Section \ref{subsec:lt}, and the proposed deep multi-view models are presented in Section \ref{subsec:arc}.
\begin{figure}[b]
\centering
\includegraphics[width=0.5\textwidth]{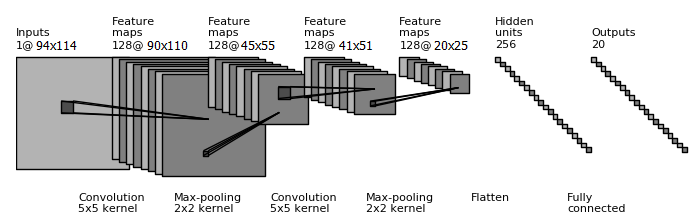}
\caption{Architecture of the ``merged view'' model.}
\label{fig:mview}
\end{figure}
\begin{table}[h!]
\centering
\caption{Best Models Specifications}
\label{tab:spec}
\begin{adjustbox}{max width=0.5\textwidth}
\begin{tabular}{l|l|l}
\hline
Single view models  & Parallel view model & Merged view model  \\ \hline
Input $1 \times 47 \times 57$ & Input four $1 \times 47 \times 57$  & Input $1 \times 94 \times 114$ \\ \hline
\begin{tabular}[c]{@{}l@{}}$5 \times 5$ Conv- $128$,\\ $2 \times 2$ Maxpooling,\\ ReLU\end{tabular} & \begin{tabular}[c]{@{}l@{}}Four $5 \times 5$ Conv- $128$,\\ $2 \times 2$ Maxpooling,\\ ReLU\end{tabular} & \begin{tabular}[c]{@{}l@{}}$5 \times 5$ Conv- $128$,\\ $2 \times 2$ Maxpooling,\\ ReLU\end{tabular} \\ \hline
\begin{tabular}[c]{@{}l@{}}$5 \times 5$ Conv- $128$,\\ $2 \times 2$ Maxpooling,\\ ReLU\end{tabular} & \begin{tabular}[c]{@{}l@{}}$5 \times 5$ Conv- $128$,\\ $2 \times 2$ Maxpooling,\\ ReLU\end{tabular} & \begin{tabular}[c]{@{}l@{}}$5 \times 5$ Conv- $128$,\\ $2 \times 2$ Maxpooling,\\ ReLU\end{tabular} \\
Fully Connected-$256$  & Fully Connected-$256$ & Fully Connected-$256$   \\
Sofmax-$20$    &  Sofmax-$20$& Sofmax-$20$  \\
  &     &                                                
\end{tabular}
\end{adjustbox}
\end{table}
%\subsection{Architectures}
The main motivation for this study is to exploit multiple views for glitch classification instead of depending on just a single view. We investigate this by combining views' information at two points as we go through the deep network layers.  We thus propose one model in which fusion take place at an early step (referred to as ``merged view'') and one in which information is integrated at the middle level (referred to as ``parallel view''). In the following sections we explain these two architectures in detail. %In the following, two primary models that consider fusion at early and late steps are presented.
%The main motivation for this study is to exploit multiple views for glitch classification instead of depending just on one view. We investigate fusion of different views at a couple of points through the deep network layers {\color{red} ye kam in jomle namafhume}. In the following, two primary models that consider fusion at early and late step are presented.
\label{subsec:arc}
%\begin{itemize}
\subsection{Parallel view model} 
The idea behind the parallel view model is to project each view into a feature space which is based more on the statistical properties of the samples than their view-specific properties. This projection makes the views interact with each other and be presented in a common feature space more efficiently. We illustrate the construction of the parallel view model using the four durations as views in Fig.~\ref{fig:pview}.
At the very first layer, each view passes through a convolutional layer, followed by max-pooling and ReLU activation. Then, we introduce a shared layer (merger layer) to map all views into a common feature space. Another set of convolutional, max-pooling, and activation layers is used after the merger layer to model the obtained common features from the previous layers. In the end, a fully connected layer and a softmax layer are employed (see Fig.~\ref{fig:pview}).
%\vspace{-1 cm}
\subsection{Merged view model}
In this model, we introduce the network layers on top of the merged views. We merge the views by forming a $2m \times 2k$ matrix by placing next to each other four $m \times k$ images. After merging the views, a set of convolutional layer followed by max-pooling and activation layer is used, and then a fully connected layer and a softmax layer are exploited (see Fig.~\ref{fig:mview}).
This approach clearly models jointly the distribution of the different views in their original feature space. This seems a reasonable approach since the correlations among the views in our problem are not highly non-linear compared to other tasks where the views are very different, such as image and text \cite{srivastava2012multimodal}, or audio and video \cite{ngiam2011multimodal}. Therefore it is possible for the model to learn such correlations even in the original space of data.
\subsection{Training}
The models presented above optimize a loss function defined on the training data. For training the model, we can use either the mean squared error or the average cross-entropy error as the loss function. Due to the advantages of the average cross-entropy error over the mean squared error, e.g., the derivation of a ``better'' gradient for back propagation, for multi-class classification problems \cite{golik2013cross}, in our model we use a cross-entropy based loss function defined as follows: 
\begin{equation}
E = -\sum_{n=1}^{N} \sum_{i=1}^{C} y_i^n \log o_i^n
\end{equation}
\noindent
where $o_i^n$ is the model's output for class $i$ when the $n^{\textrm{\textit{th}}}$ training sample is given to the network, $y_i^n$ is one if the $n^{\textrm{\textit{th}}}$ sample is from class $i$, otherwise it is zero, and $N$ and $C$ are the total numbers of the training samples and classes, respectively. There exits many optimization techniques \cite{zeiler2012adadelta,opt,vahid1,vahid2} that we can use to optimize the objective function. We use the Adadelta \cite{zeiler2012adadelta} optimizer. It monotonically decreases the learning rate and shows good performance in our experiments. 

\section{EXPERIMENTS AND RESULTS}
\label{sec:exp}
%\vspace{-0.2 cm}
\subsection{Dataset}The dataset consists of glitch samples from $20$ classes. The glitches are represented as a spectrogram, a time-frequency representation where the x-axis represents the duration of the glitch and the y-axis shows the frequency content of the glitch. The colors indicate the ``loudness'' of the glitch in the aLIGO detector. The classes arise from different environmental and instrumental mechanisms, and are based on the shape and intensity of the glitch in time-frequency space. 
The primary sample duration is $0.5$ second. However, in this study we use three other durations, i.e., $1.0$, $2.0$ and $4$ seconds, per glitch to train our multi-view models. An example of the dataset with four durations is shown in the Fig.~\ref{fig:ex}. In total, there are $7730$ glitches in our dataset. We use $75\%$, $12.5\%$, and, $12.5\%$ of samples as training, validation, and test sets, respectively\footnote{This dataset will be publicly available soon.}. 
\begin{figure*}[t!]
\centering
\includegraphics[width=0.8\textwidth]{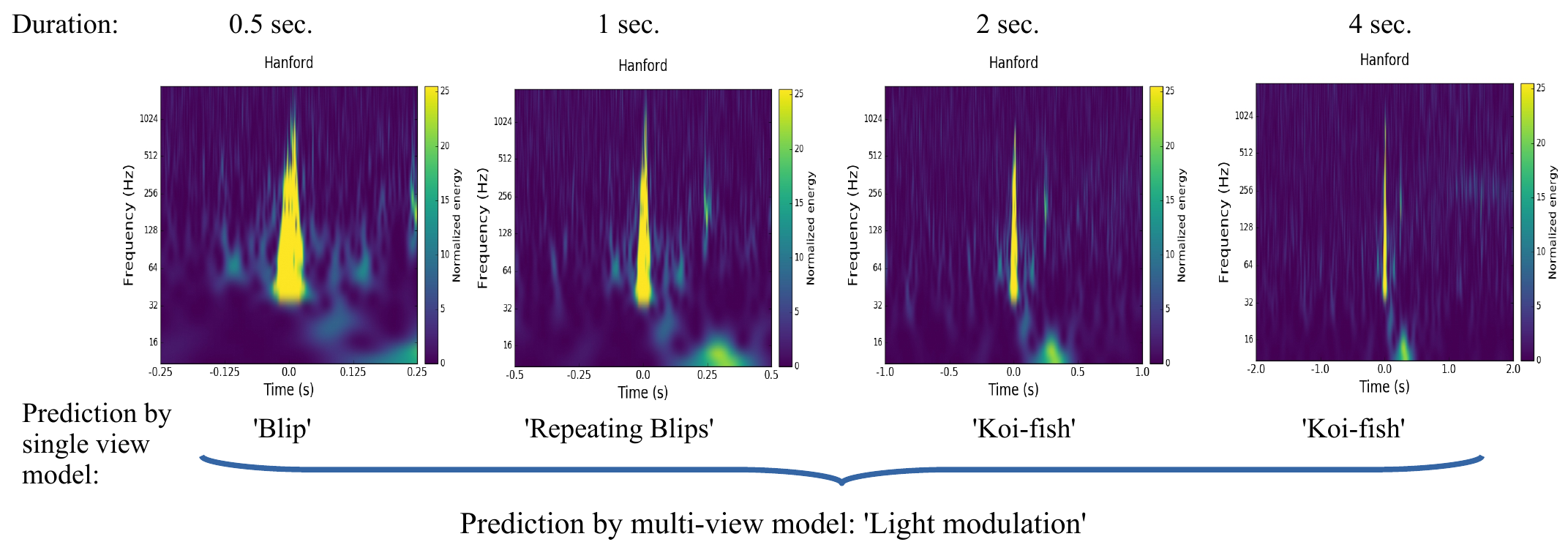}
\caption{An example of a glitch that was misclassified by all four single view models, but correctly classified with both of the multi-view models. The single view model, which is trained with $0.5$ second duration images, classifies it as ``Blip'' class. The predicted class is ``Repeating Blips'', ``Koi fish'', and ``Koi fish'' for the single view models trained with $1$, $2$, and $4$ second duration glitches, respectively. Multi-view models predict the sample correctly as belongs to the ``Light Modulation'' class. }
\label{fig:mis}
\end{figure*}
\subsection{Experiment}
\noindent
\textbf{Baseline}\\
A straightforward approach is to use just one glitch duration, as is done in a traditional single view approach. We use this as a baseline to compare the performance of our multi-view deep models. For single view models, we use CNNs with the structure shown in Table \ref{tab:spec} (left column). The architecture of CNNs is optimized for the best classification accuracy. We use two convolutional layers with $128$ kernels of size $5 \times 5$, max-pooling of $2 \times 2$, and the ReLU activation function. Batch size is set to $30$, and the number of iterations is $130$. In Table \ref{tab:sin1}, the classification accuracies of single view CNNs are presented.
\begin{table}[h!]
\centering
\caption{Classification accuracy of single view CNNs. The best model shows $95.34\%$ obtained from the model trained with $1$ second glitches. It seems that Classifier $2$ can capture the required information of both long and short duration classes better compared to the Classifier $1$ (trained with short durations) and Classifier $3$ and $4$ (trained with long durations).}

\label{tab:sin1}
\begin{tabular}{l|l|l}
Classifier & Duration & Accuracy ($\%$) \\ \hline
Classifier $1$ & $0.5$ second & 92.85 \\
Classifier $2$ & $1$ second & \textbf{95.34} \\
Classifier $3$ & $2$ seconds & 94.09 \\
Classifier $4$ & $4$ seconds & 93.68     
\end{tabular}
\end{table}

\noindent
\textbf{Deep multi-view models}\\
In the parallel view model, first, we use four separate convolutional layers (see Fig.~\ref{fig:pview}). Each has $128$ kernels with size $5 \times 5$, and $2\times 2$ max-pooling and ReLu activation. Then, we merge the output of these four convolutional layers into the merger layer and use another convolutional layer with the same structure, and finally a fully connected layer with $256$ nodes and a softmax layer with $20$ nodes (equal to the number of classes). All these details are shown in the middle column of Table \ref{tab:spec}. The parameters of this architecture were obtained based on extensive experimentation and guidance from literature.

In the merged view model, we use the structure shown in Fig.~\ref{fig:mview}. Two convolutional layers with $128$ kernel of size $5\times 5$, max-pooling of $2\times2$, and ReLU activation function are used. One fully connected layer with $256$ nodes and softmax with $20$ outputs are added to the model. All details are shown in the right column of Table \ref{tab:spec}.

For both structures, the number of iterations is set to $130$ and the batch size is $30$.  We use Keras \cite{chollet2015} with Theano \cite{theano} back-end for all implementations.
In Table \ref{tab:sin}, the best performances of parallel and merged view models are compared with the best single view model performance. See Table \ref{tab:spec} for full models specifications. 
%\vspace{-0.5 cm}
 \begin{table}[t!]
\centering
\caption{Classification accuracy of single and multi-view CNNs.}
\label{tab:sin}
\begin{tabular}{l|l}
Classifier     & Accuracy ($\%$) \\ \hline
The best single view model & 95.34 \\
parallel view model & 95.75 \\
merged view model & \textbf{96.89}    
\end{tabular}
\end{table}
\vspace{-3mm}
\begin{table}[b!]
\centering
\caption{Classification accuracy of single view model trained with one second duration (classifier $1$) and four second duration (classifier $4$) for each class. The first column shows whether the class is a short (Sh) or long (L) duration.}
\label{tab:accperclass}
 \begin{adjustbox}{max width=0.495\textwidth}
\begin{tabular}{l|l|l|l}
Duration & Class & Classifier 1 Acc. ($\%$) & Classifier 4 Acc. ($\%$)\\ \hline
Sh & Air Compressor & 100.00 & 80.00 \\  
Sh & Blip & 98.09 & 97.61 \\ 
Sh & Helix & 96.66 & 96.66  \\  
%Sh & Koi Fish & 96.00 &  100.00\\ 
%Sh & Paired Doves & 50.00 &  100.00 \\  
Sh & Power Line & 100.00 & 100.00 \\   
Sh & Repeating Blips & 80.00 & 76.00 \\  
%\vspace{1mm}
Sh & Tomte & 100.00 & 100.00 \\  \hdashline
%\vspace{1mm}
L & Extremely Loud & 96.77 & 98.38 \\  
L & Light Modulation & 83.14 & 86.51 \\ 
L & Low Frequency Lines & 89.79 & 89.79  \\  
L & Scattered Light & 98.36 &  98.36\\ 
%L & Scratchy & 97.67 &  93.02 \\  
L & Wandering Line & 57.14 & 71.42 \\     
\end{tabular}
\end{adjustbox}
\end{table}
\subsection{Analysis}
As the results in Table \ref{tab:sin} show, the performance of multi-view deep models is better than single view models. An example of a misclassified sample by all single view models that was classified correctly by the multi-view models is shown in Fig.~\ref{fig:mis}. Such examples show that in many cases, single view models do not have the needed sight and horizon for recognizing glitches correctly. 
%All the glitches in our pre-processing step are set to happen at time zero. Although the model trained with limited duration (such as $0.5$ and $1$ second) can recognize glitches in many times, there are glitches that need both the information of the zero zone and also what happens sometime before and further. That is the main reason that multi-view models show classify glitches more accurately.
Glitch classes are divided into short and long duration based on the glitch duration. In Table \ref{tab:accperclass}, we show the category of each class plus the accuracy of two of the single view models; the 0.5-second model (Classifier $1$) and 4-second model (Classifier $4$). As can be seen in this table Classifier $1$ performs at least as good as Classifier $4$ for short duration glitches, while the opposite is true for long duration glitches, as expected for some classes (e.g., ``Air Compressor'' and ``Tomte'') the performance is perfect with both classifiers. Clearly the multi-view models which use all durations can capture the needed information to classify all types of glitches (according to their duration) more accurately.
%\vspace{-0.3cm}

\section{CONCLUSIONS}
\label{sec:conc}
In this paper, we proposed multi-view deep neural network models for the glitch classification problem in aLIGO data. Two multi-view models, merged view and parallel view, are presented. The parallel view model projects samples into a common feature space where the views can interact efficiently. In the merged view model, the deep model is introduced on the concatenated durations. The experimental results show that multi-view models provide higher classification accuracy compared to the single view models, since they can accommodate efficiently the various classes independently of the glitch durations.
\section{ACKNOWLEDGMENT}
This work was supported in part by an NSF INSPIRE grant (award number IIS-1547880). The authors would like to thank Joshua Smith from California State - Fullerton University for being the chief point of contact between this study and the LIGO detector characterization working group and for being a resource on LIGO glitches and current methods of data analysis.

\bibliographystyle{IEEEbib}
\bibliography{glitch.bib}

\end{document}